\documentclass[9pt]{article}
\usepackage{spconf,amsmath,graphicx}
 \usepackage[ruled,vlined,linesnumbered]{algorithm2e}
 \usepackage{amsthm}
\usepackage{lineno}
\usepackage{color, soul}
\usepackage{amsmath,amssymb,amsfonts}
\usepackage{graphicx}
\usepackage{subfigure}
\usepackage{enumerate}
\usepackage{algpseudocode} 
\usepackage{textcomp}
\usepackage{xcolor}
\usepackage{makecell}
\usepackage{bm}
\usepackage{amsmath} 
\usepackage{float}
\usepackage{color, soul}
\usepackage{setspace}
\usepackage{lineno,hyperref} 
\usepackage{cite}
\usepackage{algpseudocode} 
\usepackage{amsmath,amssymb,amsfonts}
\usepackage{graphicx}
\usepackage{textcomp}
\usepackage{xcolor}

\def\BibTeX{{\rm B\kern-.05em{\sc i\kern-.025em b}\kern-.08em
    T\kern-.1667em\lower.7ex\hbox{E}\kern-.125emX}}
    
\title{
Adaptive Incentive for Cross-silo Federated Learning: A Multi-agent Reinforcement Learning Approach
}
\name{
Shijing Yuan$^{1}$  Hongze Liu$^{1}$  Hongtao Lv$^{2}$ Zhanbo Feng$^{1}$ Jie Li$^{1\star}$ Hongyang Chen$^{3}$ Chentao Wu$^{1}$\thanks{$^{\ast}$Corresponding author.}
}
\address{
$^{1}$Department of Computer Science and Engineering, MoE Key Lab of Artificial Intelligence,\\ AI Institute, Shanghai Jiao Tong University, China \\
$^{2}$School of Software, Shandong University, Jinan, China\\
$^{3}$Research Center for Graph Computing,
 Zhejiang Lab, China}

\begin{document}
\ninept 

\maketitle

\begin{abstract}
Cross-silo federated learning (FL) is a typical FL that enables organizations (e.g., financial or medical entities) to train global models on isolated data. Reasonable incentive is key to encouraging organizations to contribute data. {However, existing works on incentivizing cross-silo FL lack consideration of the environmental dynamics (e.g., precision of the trained global model and data owned by uncertain clients during the training processes). Moreover, most of them assume that organizations share private information, which is unrealistic.} To overcome these limitations, we propose a novel adaptive mechanism for cross-silo FL, towards incentivizing organizations to contribute data to maximize their long-term payoffs in a real dynamic training environment. The mechanism is based on multi-agent reinforcement learning, which learns near-optimal data contribution strategy from the history of potential games without organizations’ private information. Experiments demonstrate that our mechanism achieves adaptive incentive and effectively improves the long-term payoffs for organizations. 
\end{abstract}

\section{Introduction}
Federated learning (FL) is a distributed learning paradigm that enables clients to train global models distributed on isolated data. Cross-silo FL is a typical FL in which the clients consist of different organizations (e.g., financial and pharmaceutical companies) \cite{TKDE_FL}.
In cross-silo FL, organizations contribute data to train the global model (e.g., long-term cross-silo FL processes such as drug discovery and vaccine development), where local data owned by organizations may vary dynamically \cite{TMC22zhang}. However, organizations may be reluctant to participate in training. This is because organizations' commercial share may be compromised by their potential competitors who can also profit from the global model \cite{zhan2021incentive}. Therefore, a reasonable incentive mechanism is important and essential to encourage clients to contribute data resources to training \cite{zhan2021_TETC}.

Pertinent studies about incentivizing cross-silo FL focus on contribution evaluation of clients \cite{AAAI21xue,WiOpt21Lv}, fairness \cite{AIES20_hanyu,TPDS2021}, personalization incentivizing \cite{arxiv22_fair}, privacy overhead \cite{JSAC21sun,JSAC21ding}, and social welfare maximization \cite{TMC22zhang,ICASSP22_social_welfare,INFO21tang}. However, the above works either ignore the fact that data resources contributed by organizations may change dynamically, or rely on a functional form (the precision function) between the precision of the trained global model and the amount of data contribution. Nevertheless, the precision function is unknown in practice and varies with time \cite{ICML20acc_bound}.
In addition, the above works assume that organizations involved in training share their private information with other organizations, such as profitability, and training overhead, which is unrealistic in real cross-silo FL scenarios.

To overcome the above limitations, we design a novel multi-agent reinforcement learning (MARL)-based adaptive mechanism for cross-silo FL to achieve adaptive incentive in a real dynamic training environment. Specifically, we propose an adaptive mechanism based on payoff redistribution to encourage organizations to contribute data adaptively. 
Importantly, the calculation of the organizations' training payoffs requires only the value of the precision function instead of its specific form.
Under the proposed mechanism, we demonstrate that organizations' interaction to maximize their personal payoffs is a weighted potential game. To determine the strategy of data contribution that maximizes the long-term payoffs, we formulate the interaction among organizations that contribute data to maximize personal payoffs without knowing other organizations' private information as a multi-agent Partial Observable Markov Decision Process (POMDP). Then, we propose a \ul{M}ARL-based decentralized method that incorporates \ul{P}olicy \ul{G}radients and \ul{D}ifferential neural computer, called MPGD, which learns near-optimal data contribution strategies in a dynamic training environment from historical potential games without any other organizations' private information.

Our main contributions are summarized as follows.
\begin{itemize}
    \item \emph{To the best of our knowledge, we are the first to design an adaptive incentive mechanism for cross-silo FL in a real dynamic training environment}. In particular, our mechanism is conducive to practical adoption since a) it does not require any organizations' private information; b) it does not rely on any specific functional form of the precision of the trained global model and data contribution.
    \item We propose a MARL-based decentralized algorithm for incentivizing cross-silo FL, i.e., MPGD, that encourages organizations to decide their data contribution distributedly in a real and dynamic training environments.
    \item Experiments based on a near real-world platform and real datasets demonstrate that our mechanism can achieve adaptive incentive in a real dynamic training environment and effectively improve the long-term payoffs for organizations.
\end{itemize}
\section{Incentive Mechanism}
In this section, we propose an adaptive incentive mechanism based on payoff redistribution and prove that the interaction among organizations to maximize their payoffs is a weighted potential game.
\begin{figure}[!t]
	\centering
	\includegraphics[width=0.48\textwidth]{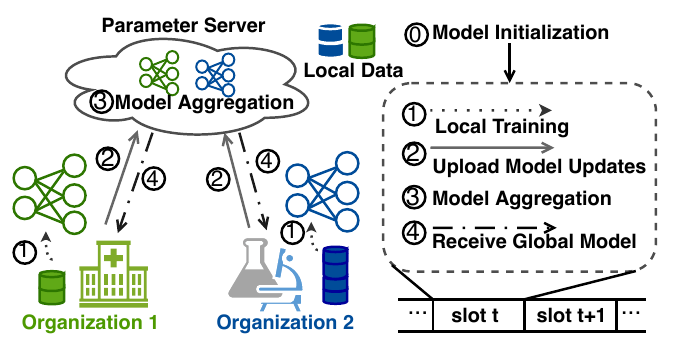}
	\vspace{-10pt}
	\caption{\scriptsize{Procedure of long-term cross-silo FL, in which organizations contribute data dynamically to maximize their personal payoffs}.}
	\label{fig:system_framework}
	\vspace{-10pt}
\end{figure}

As shown in Fig. \ref{fig:system_framework}, we consider a long-term cross-silo FL scenario with a set of $\mathcal{O} = \{o_n\}_{n\in\mathcal{N}}$ organizations, where $\mathcal{N} = \{1,\dots, N\}$. With the assistance of the parameter server, each organization $o_n\in\mathcal{O}$ contributes its local data to participate in the long-term training procedure in expectation of the trained global model. The training process is discretized into time slots $\mathcal{I} = \{0,1,\dots,T-1\}$. In each time slot $t$, $o_n\in\mathcal{O}$ contributes $x^{[t]}_{n}$ data to perform local training, where $x^{[t]}_{n}\in[0,|\mathcal{D}^{[t]}_{n}|]$, $\mathcal{D}^{[t]}_{n}$ and $|\mathcal{D}^{[t]}_{n}|$ are $o_n$'s local data set and the size of data set (the number of data samples), respectively. After the local training, $o_n$ uploads the updated model to the parameter server for model aggregation. 

To characterize the payoffs of organizations that participate in training, we quantify their revenue from the global model and training overhead in each time slot.\\
\noindent 1) \textbf{Revenue from the global model.} Let $d_n=\frac{x_n}{|\mathcal{D}_{n}|}$ be $o_n$'s data contribution strategy, and denote $\boldsymbol{d}^{[t]}_{-n} = \{d^{[t]}_{j}\}_{j\in\mathcal{N},j\neq n}$ as that of all organizations except $o_n$ within the $t^{th}$ time slot. The precision of the trained global model after the $t^{th}$ time slot can be denoted as $P^{[t]}(d^{[t]}_{n},\boldsymbol{d}^{[t]}_{-n})$. Note that $P^{[t]}(d^{[t]}_{n},\boldsymbol{d}^{[t]}_{-n})$\footnote{In this paper, the precision of the trained global model $P^{[t]}(d^{[t]}_{n},\boldsymbol{d}^{[t]}_{-n})$ is called \emph{precision function}. According to \cite{TMC22zhang,ICASSP22_social_welfare,INFO21tang}, $P^{[t]}(d^{[t]}_{n},\boldsymbol{d}^{[t]}_{-n})$ is a convex function with a strongly convex loss function. That is, given $\boldsymbol{d}_{-n}$, $P(d_{n},\boldsymbol{d}_{-n})$ improves as the amount of data invested by $o_n$ increases, but the growth rate decreases. Therefore, we assume that $P(d_{n},\boldsymbol{d}_{-n})$ satisfies $\bigtriangledown_{d_{n}}P(d_{n},\boldsymbol{d}_{-n})\ge 0 $.} varies with time slot $t$ and can be observed by all organizations at the end of the $t^{th}$ time slot. Let $p_n$ be the profit obtained by $o_n$ from unit performance of the global model, and thus the revenue obtained by $o_n$ from the global model can be denoted as $p_nP^{[t]}(d^{[t]}_{n},\boldsymbol{d}^{[t]}_{-n})$ \cite{TMC22zhang,INFO21tang}.\\

\noindent 2) \textbf{Training overhead.} In the $t^{th}$ time slot, $o_n$'s training overhead consists of the energy consumption $E^{[t]}_{n}$ and the communication overhead $C^{[t]}_{n}$, where $E^{[t]}_{n}= v_n d^{[t]}_{n}|\mathcal{D}_{n}|$, and $v_n$ is the energy consumption per unit training sample of $o_n$ \cite{ICASSP21_FL_lyapunov,TC21_LSTM}.

Next, we propose an adaptive incentive mechanism based on payoff redistribution to encourage organizations to contribute their local data.
The motivation is that potential competition among organizations may prevent them from contributing their data, as organizations' competitors can also profit from the global model, and thus their commercial share can be compromised \cite{zhan2021incentive,yuhan2021_TBD}. The proposed payoff redistribution enables the organization with less data contribution to compensate the organization with more data contribution. Consequently, the payoff redistribution received by $o_n$ is $    r^{[t]}_{n}(d^{[t]}_{n},\boldsymbol{d}^{[t]}_{-n}) = \sum_{j\in \mathcal{N}} \alpha^{[t]}(d^{[t]}_n -d^{[t]}_j)$,
where $\alpha^{[t]}$ represents the intensity of payoff redistribution. Clearly, $r^{[t]}_{n}$ varies adaptively with the data contribution $\{d^{[t]}_{n},\boldsymbol{d}^{[t]}_{-n}\}$.

\noindent3) \textbf{Payoff of participating in training}. Finally, $o_n$'s payoff at the $t^{th}$ time slot can be formulated as
\vspace{-5pt}
\begin{equation}\label{eq:utility}
    u^{[t]}_{n}(d^{[t]}_{n},\boldsymbol{d}^{[t]}_{-n}) = p_nP^{[t]}(d^{[t]}_{n},\boldsymbol{d}^{[t]}_{-n}) -  v_n d^{[t]}_{n}|\mathcal{D}_{n}| -C^{[t]}_{n} +r^{[t]}_{n}.
    \vspace{-5pt}
\end{equation}
Thus, the calculation of Eq. (\ref{eq:utility}) requires only the value of $P^{[t]}$ without its specific form.

With the above mechanism, we apply the game theory \cite{2016potential_game_theory} to analyze the optimal data contribution strategy for each organization, since the game theory is a powerful method for analyzing interactions among multiple clients. In detail, the interaction among organizations that pursue payoff maximization can be formulated as a non-cooperative game $\boldsymbol{\mathcal{G}}=\{\mathcal{N},\{d^{[t]}_{n}\}_{n\in \mathcal{N}},\{u^{[t]}_{n}\}_{n\in \mathcal{N}}\}$, where each organization $o_n$ makes a decision $d^{[t]}_{n}$ to maximize its' payoff with given $\boldsymbol{d}^{[t]}_{-n}$
\vspace{-3pt}
\begin{equation}\label{prob:game}
    \begin{aligned}
        & \max_{d^{[t]}_{n}} \quad  u^{[t]}_{n}(d^{[t]}_{n},\boldsymbol{d}^{[t]}_{-n}) \\
        &\begin{aligned}
          \text { s.t. } d^{[t]}_{n}\in[0,1].\\ 
        \end{aligned}
    \end{aligned}
    \vspace{-8pt}
\end{equation}


Accordingly, the objective of Eq. (\ref{prob:game}) is to achieve $\boldsymbol{\mathcal{G}}$'s Nash equilibrium (NE), whose concept is defined as follows. 
\noindent\textbf{Definition 1} \textbf{(Nash Equilibrium)}.
\emph{Nash equilibrium (NE) of $\mathcal{G}$ is a strategy profile $ \boldsymbol{d^{[t]*}} = \{ d^{[t]*}_{n}\}_{n \in \mathcal{N}}$ in which no $o_n$ can further increase its payoff by unilaterally changing its strategy}
\vspace{-5pt}
\begin{equation} \label{eq:NE}
u_n(d^{[t]*}_{n},\boldsymbol{d}^{[t]*}_{-n})  \ge u_n(d^{[t]*}_{n},\boldsymbol{d}^{[t]*}_{-n}) , 
     \quad \forall n \in \mathcal{N}.
\end{equation}

\noindent To prove the existence of NE, we introduce the definition of weighted potential game as follows.

\begin{spacing}{0.95}
\noindent\textbf{Definition 2} \textbf{(Weighted Potential Game)}. \emph{A game $\boldsymbol{\mathcal{G}}$ is a weighted potential game \cite{2016potential_game_theory}, if there is a potential function  $\boldsymbol{U}(d^{[t]}_{n},\boldsymbol{d}^{[t]}_{-n})$ such that the change in each $o_n$'s payoff due to its strategy deviation is equal to the change in the potential function but scaled by a non-negative weighting factor $w_n$, i.e.,}
\end{spacing}
\vspace{-5pt}
\vspace{-5pt}
\begin{small}
\vspace{-5pt}
\begin{equation}\label{eq:potential_function}
\vspace{-5pt}
 \begin{aligned}
    w_n \left[ \boldsymbol{U}(\boldsymbol{d}^{[t]\prime}) - \boldsymbol{U}(\boldsymbol{d}^{[t]}) \right] =  
    {u}_{n}(\boldsymbol{d}^{[t]\prime}) - {u}_{n}(\boldsymbol{d}^{[t]}),
   \end{aligned}
\end{equation}  
\end{small}%
\begin{spacing}{0.95}
\noindent where $\boldsymbol{d}^{[t]} = \{d^{[t]}_{n},\boldsymbol{d}^{[t]}_{-n}\}$ and $\boldsymbol{d}^{[t]\prime} = \{d^{[t]\prime}_{n},\boldsymbol{d}^{[t]}_{-n}\}$.
Note that NE is guaranteed to be present in every potential game.
\end{spacing}

\noindent \textbf{Theorem 1 ($\boldsymbol{\mathcal{G}}$ is A Weighted Potential Game).} \emph{ $\boldsymbol{\mathcal{G}}$ is a weighted potential game with the potential function that is given in Eq. (\ref{eq:thereoum1})}
\vspace{-5pt}
\vspace{-5pt}
\begin{small}
\begin{equation}\label{eq:thereoum1}
\vspace{-10pt}
 \begin{aligned}
                     \boldsymbol{U}(\boldsymbol{d}^{[t]}) 
        =\bm{P}\left(\boldsymbol{d}^{[t]} \right)-\sum_{n\in\mathcal{N}}\left[\frac{ v_n d^{[t]}_{n}|\mathcal{D}_{n}|+C^{[t]}_{n}}{p_n}-\sum_{j\in \boldsymbol{-n}}\frac{ r^{[t]}_{n,j}}{ p_n}\right].
   \end{aligned}
\end{equation}  
\end{small}
\vspace{3pt}
\begin{spacing}{0.95}
\noindent \emph{\textbf{Proof}}.
Here, we omit the superscript $[t]$ for concision. Let the decision of any $o_n$ change from $d_n$ to $d^{\prime}_{n}$, we have 
\end{spacing}
\begin{small}
\vspace{-16pt}
\begin{equation}\label{eq:U_prime}
        \begin{aligned}
    &\boldsymbol{U}({d}^{\prime}_{n},\boldsymbol{d}_{-n})
        =\bm{P}\left(d^{\prime}_{n},\boldsymbol{d}_{-n}\right)-\sum_{j\in\mathcal{N}, j\neq n}\frac{ v_j d_{j}|\mathcal{D}_{n}|+C_{j}}{p_j}\\
        &-\frac{ v_n d^{\prime}_{n}|\mathcal{D}_{n}|+C_{n}}{p_n}+\sum_{j\in \mathcal{N}}\frac{r_{n,j}(d^{\prime}_{n},d_j)}{p_n} +\sum_{j\in\mathcal{N}, \atop j\neq i}\sum_{k\in \mathcal{N}}\frac{r_{j,k}(d_j,d_k)}{p_n}.\\
    \end{aligned}
    \vspace{-5pt}
     \vspace{-5pt}
\end{equation}
\end{small}
\noindent Subtracting Eq. (\ref{eq:U_prime}) from Eq. (\ref{eq:thereoum1}), we obtain
\begin{small}
\begin{equation}\label{eq:proof_theoreum1}
 \begin{aligned}
     p_n \left[ \boldsymbol{U}(d_n,\boldsymbol{d}_{-n}) - \boldsymbol{U}(d^{\prime}_{n},\boldsymbol{d}_{-n}) \right] =  
    {u}_{n}(d_n,\boldsymbol{d}_{-n}) - {u}_{n}(d^{\prime}_{n},\boldsymbol{d}_{-n}),
   \end{aligned}
\end{equation}  
hence $\mathcal{G}$ is a weighted potential game that the NE exists.
\end{small}$\qedsymbol$

\begin{spacing}{0.95}
Although $\mathcal{G}$ is proved to be a weighted potential game, the NE of $\mathcal{G}$ can not be achieved by traditional methods (e.g., dynamic response based algorithm \cite{SIAM_best_response2018}) for the following reasons.
\emph{1) In practical scenarios, organizations may not know private information such as other organizations' profitability $p_n$, communication overhead $C^{[t]}_{n}$, and training overhead $E^{[t]}_{n}$.
2) The functional relationship between the precision of the trained global model $P^{[t]}(d^{[t]}_{n},\boldsymbol{d}^{[t]}_{-n})$ and data contribution $\{d^{[t]}_{n},\boldsymbol{d}^{[t]}_{-n}$\}, is a “black box" that is difficult to describe quantitatively \cite{INFO21tang}}.
Based on the above observations, in Section \ref{sec:MPGD}, we formulate the long-term interaction among organizations as a multi-agent partially observable Markov decision process (POMDP), and then propose a multi-agent reinforcement learning based decentralized algorithm, i.e., \textbf{MPGD}. 
\end{spacing}

\begin{figure}[!t]
	\centering
	\includegraphics[width=0.48\textwidth]{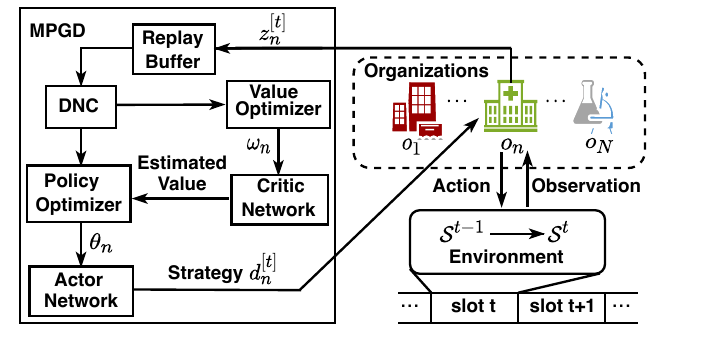}
	\vspace{-10pt}
	\caption{\scriptsize{Architecture of the proposed MPGD algorithm that enables organizations to contribute data distributedly in a dynamic training environment without the private information of other organizations}.}
	\label{fig:MPGD}
	\vspace{-10pt}
\end{figure}
\vspace{-5pt}
\section{Multi-agent Reinforcement Learning based Algorithm}\label{sec:MPGD}

\vspace{-5pt}
\subsection{Partially Observable Markov Decision Process}
\vspace{-5pt}
 We formulate the process of organizational contributing data to maximize long-term payoffs as a multi-agent partially observable Markov decision process, i.e., $\mathcal{M} = < \mathcal{S},\mathcal{A},\mathcal{P},\mathcal{Z},\mathcal{R},\mathcal{T}>$, where $\mathcal{S} = \{\mathcal{S}_{n}=\{P^{[t]},{C}_{n}^{[t]}, {E}_{n}^{[t]},$\\
 $\alpha^{[t]}\}_{t\in\mathcal{I}} \}_{n\in\mathcal{N}}$ represents the state space, $\mathcal{A} = \{\mathcal{A}_{n}=$\\
 $\{d^{[t]}_{n}\}_{t\in\mathcal{I}} \}_{n\in\mathcal{N}}$ denotes the action space, $\mathcal{P} = \{ \mathcal{S}_{n} \times \mathcal{A} \times \mathcal{S}_{n} \rightarrow [0,1] \}_{n\in \mathcal{N}}$ is the set of state transition probability function, $\mathcal{Z} = \{\mathcal{Z}_{n} = \{z^{[t]}_{n}\}_{t\in\mathcal{I}} \}_{n\in \mathcal{N}}$ is the observation space,
$z^{[t]}_{n} = [h^{[t-1]}_{n},\dots,h^{[t-H]}_{n}]^{T}$ is $o_n$'s observation at the $t^{th}$ time slot, $h^{[k]}_{n} = \{\boldsymbol{d}_{-n}^{[k]}, C_{n}^{[k]}, \alpha^{[k]}, P^{[k]}\},k\in[t-1, t-H]$, $\mathcal{R} = \{ \mathcal{R}_{n} = \{u^{[t]}_{n}(d^{[t]}_{n},\boldsymbol{d}^{[t]}_{-n}) \}_{t\in\mathcal{I}} \}_{n\in\mathcal{N}}$ is the reward space, and $\mathcal{T} = \{ \{\mathcal{S}_{n} \times \mathcal{O}_{n} \rightarrow [0,1] \}_{t\in\mathcal{I}} \}_{n\in\mathcal{N}}$ represents the observation function set. It is noteworthy that $z^{[t]}_{n}$ contains the data contribution strategy of the other organizations $\boldsymbol{d}_{-n}^{[k]}$, $o_n$'s communication overhead $C_{n}^{[k]}$, the intensity of payoff redistribution $\alpha^{[k]}$, and the precision of the global model $P^{[k]}$ at previous $H$ time slots. 
 \vspace{-8pt}
\vspace{-5pt}
\subsection{Algorithm Design}
 \vspace{-2pt}
 \vspace{-5pt}
Next, we propose a multi-agent reinforcement learning based decentralized algorithm, called, \textbf{MPGD}, incorporating policy gradient methods and differential neural computer (DNC) \cite{zhan2020deep} to determine the data contribution strategy $\{\{d^{[t]}_{n}\}_{t\in\mathcal{I}}\}_{n\in\mathcal{N}}$ that maximizes the long-term payoffs for organizations in a dynamic environment. To be specific, $o_n$'s data contribution policy parameterized by $\theta_n$ is $\pi_{\theta_n}:\mathcal{Z}_{n}\times\mathcal{A}_{n}\rightarrow[0,1]$, and thus the $o_n$'s optimization objective is
\vspace{-5pt}
\begin{equation}\label{eq:Ln}
    \begin{aligned}
\theta_n^* &=\underset{\theta_n}{\arg \max } J_n\left(\pi_{\theta_n}\right) \\
&=\underset{\theta_n}{\arg \max } \mathbb{E}\left[V^{\pi_{\theta_n}}\left(\boldsymbol{z}_n^0\right) \mid \rho_n^0\right] \\
&=\underset{\theta_n}{\arg \max } \mathbb{E}\left[Q^{\pi_{\theta_n}}\left(\boldsymbol{z}_n^0, d_n^0\right) \mid \rho_n^0, \pi_{\theta_n}\right],
\end{aligned}
\vspace{-8pt}
\end{equation}
where $J_n\left(\pi_{\theta_n}\right)$ is the objective function, $V^{\pi_{\theta_n}}(\boldsymbol{z}_n)=\mathbb{E}\big[R_{n}^{t} \mid$ 
$\boldsymbol{z}_{n}^{t}=\boldsymbol{z}_n, \Pi, \mathcal{P}, \mathcal{T}\big]$
is the value function of observation, $Q^{\pi_{\theta_n}}(\boldsymbol{z}_n, d_n)=\mathbb{E}\big[R_n^t \mid \boldsymbol{z}_n^t=\boldsymbol{z}_n,$ $ d_n^t=d_n, \Pi, \mathcal{P}, \mathcal{T}\big]$ is the value function of observation and action, $\rho_n^0$ is the initial observation probability distribution of $z_n$, $\Pi=\left\{\pi_{\theta_n}\right\}_{n \in \mathcal{N}}$ denotes the set of all organizations' policies, $R_n^t=\sum_{l=t}^T \gamma^{l-t} u_n^t$ is the discounted expected future reward, and $\gamma \in[0,1]$ is the discount factor.

As shown in Fig. \ref{fig:MPGD}, \textbf{MPGD} consists of a Critic-Network parameterized by $\omega$, in which each observation is mapped into a feature vector, an Actor-Network parameterized by $\theta$ that outputs actions conditional on observation to approximate policy, a replay buffer that stores the history game records, and a DNC module, which is a recurrent neural network \cite{zhan2020deep} that helps the organizations' strategy to converge to the NE faster. To accelerate the convergence of MPGD, we optimize the Actor-Network and the Critic-Network as follows.
 \vspace{-10pt}
\subsubsection{Optimization of the Actor Network}
 \vspace{-2pt}
We employ the policy gradient theorem \cite{zhan2020deep} to optimize the actor network, which clips the policy gradient.

\noindent\textbf{Proposition 1}. \emph{The policy gradient of MPGD is}
\vspace{-3pt}
\begin{equation}\label{eq:actor_optimization}
    \begin{aligned}
\nabla_{\theta_n} J_n &=\mathbb{E}_{\pi_{\theta_n}, \rho_n^1\left(\boldsymbol{z}_n\right)}\left[\nabla_{\theta_n} \log \pi_{\theta_n}\left(\boldsymbol{z}_n, d_n\right) Q^{\pi_{\theta_n}}\left(\boldsymbol{z}_n, d_n\right)\right] \\
&=\mathbb{E}_{\pi_{\theta_n}, \rho_n^1\left(\boldsymbol{z}_n\right)}\left[\nabla_{\theta_n} \log \pi_{\theta_n}\left(\boldsymbol{z}_n, d_n\right) A^{\pi_{\theta_n}}\left(\boldsymbol{z}_n, d_n\right)\right] \\
& \approx \mathbb{E}_{\pi_{\widehat{\theta_n}}, \rho_n^1\left(\boldsymbol{z}_n\right)}\left[f_n \nabla_{\theta_n} \log \pi_{\theta_n}\left(\boldsymbol{z}_n, d_n\right) A^{\pi_{\theta_n}}\left(\boldsymbol{z}_n, d_n\right)\right],
\end{aligned}
\vspace{-7pt}
\end{equation}%
\emph{where $f_n=\frac{\pi_{\theta_n}\left(z_n \mid d_n\right)}{\pi_{\widehat{\theta_n}}\left(z_n \mid d_n\right)}, A^{\pi_{\theta_n}}\left(\boldsymbol{z}_n, d_n\right)=Q^{\pi_{\theta_n}}\left(\boldsymbol{z}_n, d_n\right)-V^{\pi_{\theta_n}}\left(\boldsymbol{z}_n\right)$,
$\hat{\theta}$ is the policy parameter for sampling, and $\rho_n^1\left(z_n\right)$ is the observation distribution.}

\noindent \emph{\textbf{Proof}}. By definition, $V^\pi(s)=\sum_d \pi(s,  d) Q^\pi(s,  d)$, we have
\vspace{-5pt}
\begin{small}
\begin{equation}
    \nabla_\theta V^\pi(s)=\sum_{s \in \mathcal{S}} d^\pi(s) \sum_d \pi(s,  d) \nabla_\theta \log \pi(s,  d) Q^\pi(s,  d),
    \vspace{-10pt}
\end{equation}
\end{small}%
\begin{spacing}{0.95}%
\noindent where $d^\pi(s)=\sum_{t=0}^{\infty} \gamma^t \operatorname{Pr}\left(s_0 \rightarrow s, t, \pi\right)$ is the discounted state distribution.
\begin{figure*}
	\setlength{\abovecaptionskip}{-2pt}
	\setlength{\belowcaptionskip}{-5pt}
	\centering
		\begin{minipage}[t]{0.32\textwidth}
		\centering
\includegraphics[width=2.06in]{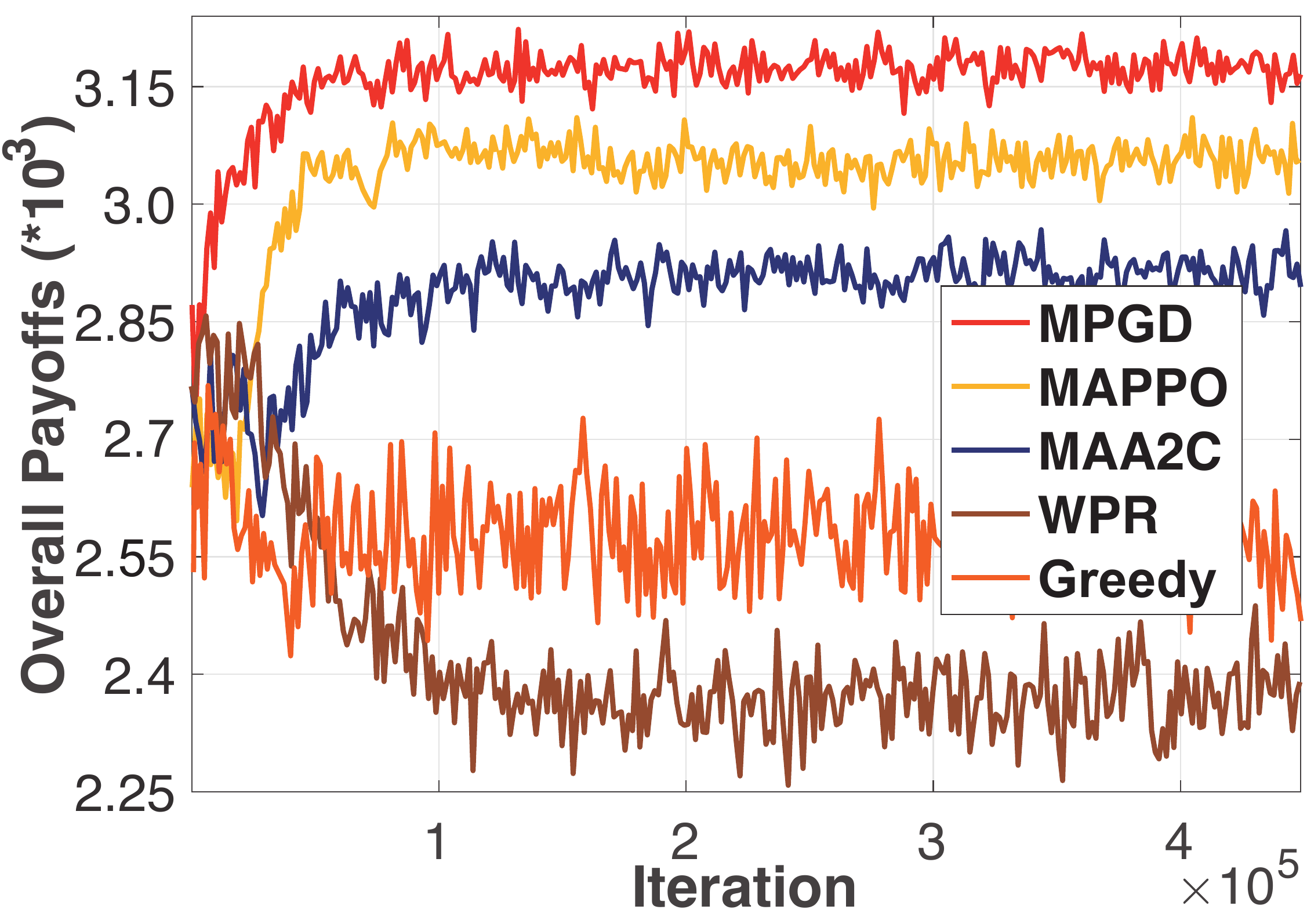}
		\vspace{-5pt}
	 \caption{\scriptsize{Overall payoffs under four different schemes.}
        }
		\label{fig:overall_utility}
	\end{minipage}
		\begin{minipage}[t]{0.32\textwidth}
		\centering
	\includegraphics[width=2.1in]{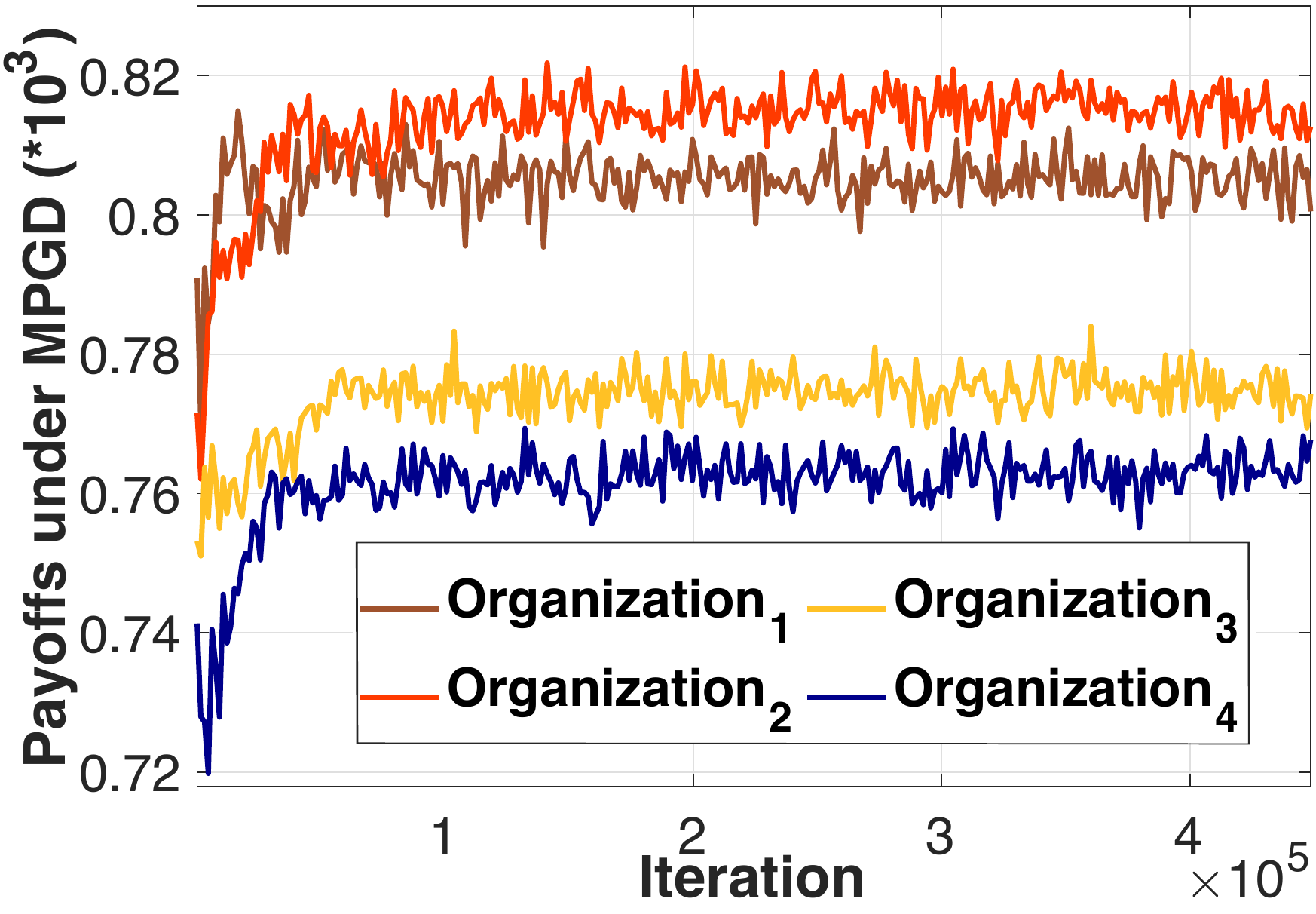}
	\vspace{-5pt}
	 \caption{\scriptsize{Each organization's payoff under MPGD.}
        }
		\label{fig:utility_MPGD}
	\end{minipage}
		\begin{minipage}[t]{0.32\textwidth}
		\centering
	\includegraphics[width=2.1in]{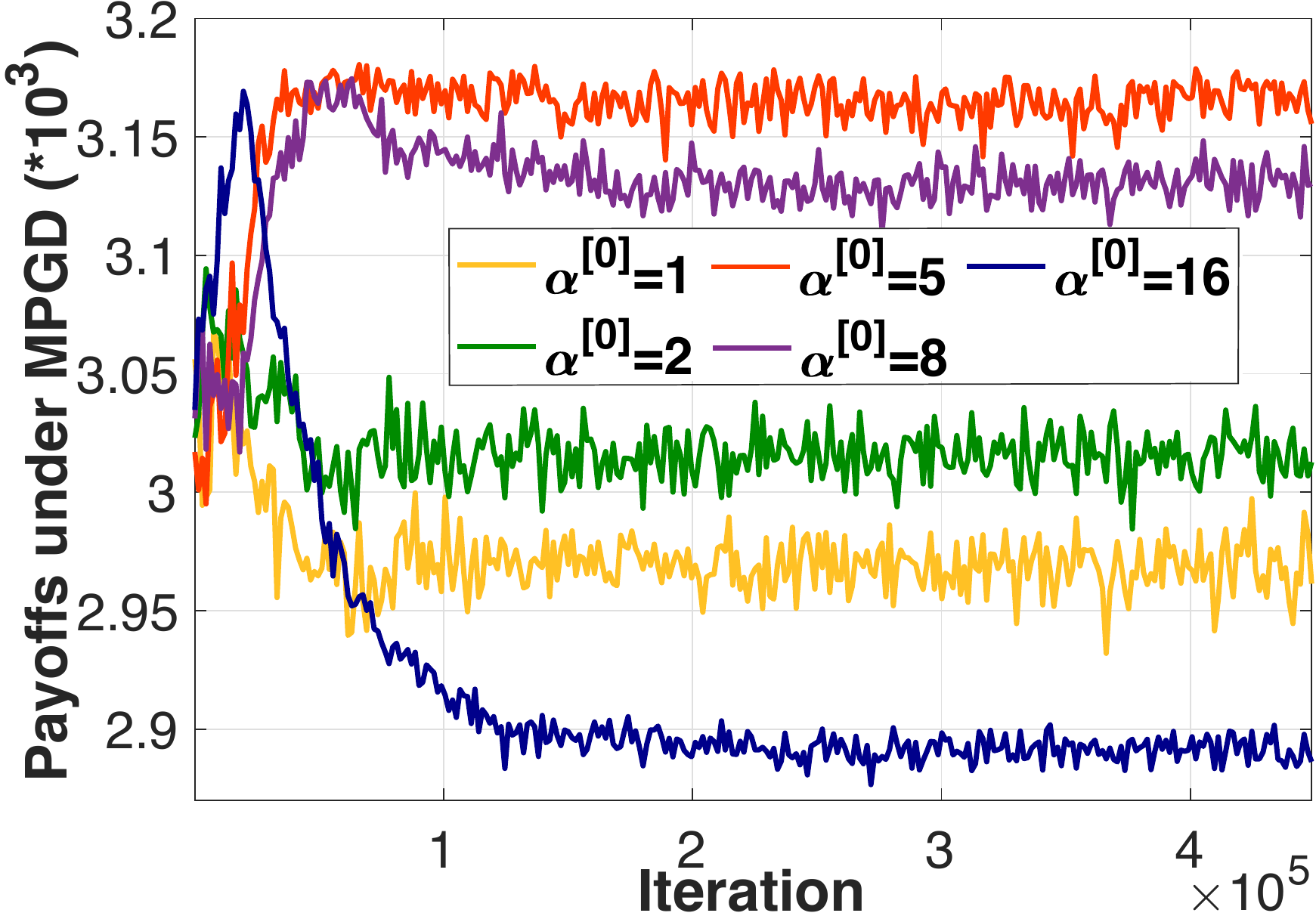}
	\vspace{-5pt}
	 \caption{\scriptsize{Impact of the intensity of payoff redistribution on organizations' payoffs.}
        }
		\label{fig:impact_alpha}
	\end{minipage}
\vspace{-5pt}
\vspace{-5pt}
\end{figure*}
Therefore, when we take the expectation over the state $s \sim(1-\gamma) d^\pi(s)$ and the action $d \sim \pi(s,  d)$, we have
\end{spacing}
\vspace{-5pt}
\vspace{-5pt}
\vspace{-5pt}
\begin{small}
\begin{equation}
    \begin{aligned}
\nabla_\theta V^\pi(s) &=\frac{1}{1-\gamma} \sum_{s \in \mathcal{S}}d^\pi_\gamma \sum_d \pi(s,  d) \nabla_\theta \log \pi(s,  d) Q^\pi(s,  d) \\
&=\frac{1}{1-\gamma} \mathbb{E}_{\pi(\theta)}\left[\nabla_\theta \log \pi(s,  d) Q^\pi(s,  d)\right],
\vspace{-5pt}
\vspace{-5pt}
\vspace{-3pt}
\vspace{-5pt}
\vspace{-5pt}
\vspace{-5pt}
\end{aligned}
\end{equation}
\end{small}%
\noindent where $d^\pi_\gamma = (1-\gamma) d^\pi(s)$. Since the observation $z$ which reflects the hidden state $s$ is sampled from the observation distribution $\rho^1(z)$, and thus we have
\begin{small}
\vspace{-5pt}
\begin{equation}
    \begin{aligned}
\nabla_\theta J=\mathbb{E}\left[V^{\pi_\theta}(z) \mid \rho\right]=\mathbb{E}_{\pi_{\theta}, \rho^1(z)}\left[\nabla_\theta \log \pi_\theta(z, d) Q^{\pi_\theta}(z, d)\right],
\vspace{-5pt}
\vspace{-5pt}
\vspace{-5pt}
\end{aligned}
\end{equation}
\end{small}%
which confirms Eq. (\ref{eq:actor_optimization}). $\qedsymbol$


Further, the policy gradient and the estimated gradient with respect to $\theta_n$ can be calculated as 
\vspace{-3pt}
\begin{equation}\label{eq:estimated_theta}
  \begin{aligned}
&\nabla_{\theta_n} J_n \approx \mathbb{E}_{\pi_{\widehat{\theta_n}}, \rho_n^1\left(\boldsymbol{z}_n\right)}\left[\nabla_{\theta_n} \log \pi_{\theta_n}\left(\boldsymbol{z}_n, d_n\right) C\left(\boldsymbol{z}_n, d_n\right)\right],\\
    &\nabla_{\theta_n} \hat{J}_n=\frac{1}{D} \sum_{t=0}^{D-1} \nabla_{\theta_n} \log \pi_{\theta_n}\left(\boldsymbol{z}_n^t, d_n^t\right) C\left(\boldsymbol{z}_n^t, d_n^t\right), \text{where} 
  \end{aligned}
  \vspace{-5pt}
  \vspace{-5pt}
\end{equation}
\vspace{-7pt}
\begin{equation} \label{eq:C}
    C\left(\boldsymbol{z}_n, d_n\right)=\min \left\{f_n A^{\pi_{\theta_n}}\left(\boldsymbol{z}_n, d_n\right), \eta\left(f_n\right) A^{\pi_{\theta_n}}\left(\boldsymbol{z}_n, d_n\right)\right\},
\end{equation}
\vspace{-15pt}
\begin{equation} \label{eq:zeta}
\begin{aligned}
    \eta(d)= \begin{cases}1+\epsilon, & d>1+\epsilon \\
d, & 1-\epsilon \leq d \leq 1+\epsilon, \\
1-\epsilon, & d<1-\epsilon\end{cases}
\end{aligned}
\end{equation}
and $\epsilon$ is an adjustable parameter. The actor network is updated by stochastic gradient descent, i.e., $\theta_n \leftarrow \theta_n+l_{a,n} \nabla_{\theta_n} \hat{J}_n$,
where $l_{a,n}$ is the learning rate and $D$ is the mini-batch size for updating the actor network.
\vspace{-5pt}
\vspace{-5pt}
\subsubsection{Optimization of the Critic Network}
\vspace{-5pt}
The objective function for the critic network is
\begin{equation}\label{eq:obj_critic_network}
F_n\left(\omega_n\right)=\mathbb{E}_{\boldsymbol{z}_{\mathrm{n}} \sim \rho_n^1\left(\boldsymbol{z}_n\right)}\left[-V_{\omega_n}\left(\boldsymbol{z}_n\right)+\mathbb{E}_{\boldsymbol{z}_n^{\prime}, d_n}\left[u+V_{\omega_n}\left(\boldsymbol{z}_n^{\prime}\right)\right]\right]^2,
\end{equation}
where $\boldsymbol{z}_n^{\prime}$ is the next time observation of $\boldsymbol{z}_n$. Similarly, the estimated gradient with respect to $\omega_n$ is calculated as:
\vspace{-5pt}
\vspace{-5pt}
\begin{equation}\label{eq:estimated_gradient_critic}
    \nabla_{\omega_n} \hat{F}_n=\frac{1}{D} \sum_{t=0}^{D-1}\left[V_{\omega_n}\left(\boldsymbol{z}_n^t\right)-Y_n^t\right] \frac{\mathrm{d} V_{\omega_n}\left(\boldsymbol{z}_n\right)}{\mathrm{d} \omega_n},
    \vspace{-5pt}
\end{equation}
where $Y_n^t=R_n^t-\gamma^{D-t} R_n(D)+\gamma^{D-t} V_{\omega_n}\left(z_n(D)\right)$, and $D$ denotes the size of mini-batch. The critic network is updated with learning rate $l_{c,n}$, i.e., $\omega_n \leftarrow \omega_n-l_{c,n} \nabla_{\omega_n} \hat{F}_n$.

\vspace{-5pt}
\section{Experiment and Discussion}
\vspace{-5pt}

\textbf{Experimental Settings}. To evaluate the long-term payoffs of organizations, we conduct experiments with $|\mathcal{N}|=4$ clients on \emph{FedScale} \cite{lai2021fedscale}, a near real-world FL platform, using FedAvg \cite{TKDE_FL} algorithm and datasets of FMNIST \cite{FMNIST} and CIFAR-10 \cite{CIFAR10}. Following \cite{TC21_LSTM,ToN20_Cn}, the profitability and communication overhead of $o_n$ is set to $p_n \thicksim N(1000,10) $ and $C^{[t]}_{n} \thicksim N(0.5,0.02)$, respectively. The number of data samples is set to $|\mathcal{D}_{n}|\thicksim N(2000, 50)$, and the energy consumption per unit sample is set to $v_n\thicksim N(4,0.2)$. To avoid high training overhead, the intensity of payoff redistribution $\alpha$ is set to decay with the gain of $P^{[t]}$ since it decreases as time slot $t$,
i.e., $\alpha^{[t]}= \frac{P^{[t]}-P^{[t-1]}}{P^{[t-1]}-P^{[t-2]}}\alpha^{[0]}$, where $\alpha^{[0]} = 5$.
Both the actor network and the critic network have two hidden fully-connected layers, each with 210 and 50 nodes, respectively.

\noindent\textbf{Baselines}. We compare MPGD with state-of-the-art methods, including \textbf{MAA2C} \cite{zhan2020deep}, a multi-agent variant of Advantage Actor-Critic (A2C) algorithm that incorporates DNC and A2C, \textbf{MAPPO} \cite{MAPPO}, an Proximal Policy Optimization (PPO) algorithm implemented for multi-agent training, \textbf{Greedy}, where each organization makes the decision of data contribution with the maximum reward in each time slot, and \textbf{WPR} \cite{TC21_LSTM}, where organizations participate in training without the proposed incentive mechanism, meaning that payoff redistribution term $r^{[t]}_{n}$ is not contained in $o_n$'s payoff $u_{n}^{[t]}$.


Fig. \ref{fig:overall_utility} illustrates the convergence and long-term payoffs under different schemes. We can observe that MPGD achieves the highest long-term payoffs and the fastest convergence compared to other baselines. Unlike WPR, the payoffs under MPGD gradually increase and converge rapidly. This result indicates that the proposed incentive mechanism based on payoff redistribution can effectively encourage organizations to provide data to participate in training, thus improving the long-term utility of organizations.


Fig. \ref{fig:utility_MPGD} shows the payoffs of each organization under MPGD. We find that the payoffs of all organizations continue to improve due to the update of their data contribution strategy, which converges after $10^5$ iterations. This demonstrates that the proposed mechanism can achieve dynamic adaptive incentives without any organization's private information and assumptions about the specific form of $P^{[t]}(d^{[t]}_{n},\boldsymbol{d}^{[t]}_{-n})$.



Fig. \ref{fig:impact_alpha} demonstrates the impact of the initial value of ${\alpha}$ on organizations' overall payoffs. We can draw an interesting conclusion: An appropriate initial value of ${\alpha}$ is beneficial to maximize the overall payoffs of organizations. As we can see from Fig. \ref{fig:impact_alpha}, increasing ${\alpha}^{[0]}$ from $1$ to $16$ does not always improve organizations' overall payoffs. The reason is that although a larger ${\alpha}^{[0]}$ can encourage organizations to contribute more data, more data also means higher training overhead.

\vspace{-5pt}
\vspace{-5pt}
\section{Conclusion}
\vspace{-5pt}
In this work, we design an adaptive incentive mechanism for cross-silo FL, MPGD, which encourages organizations to contribute data adaptively in a real dynamic training environment. MPGD is easy to apply in practice because it does not rely on any organizational private information and does not assume any exact functional form of the precision function. Experimental results demonstrate that MPGD effectively improves the long-term payoffs over typical baselines. 
\bibliographystyle{IEEEbib}
\bibliography{IEEEabrv,ref}

\end{document}